\documentclass{article} 
\usepackage{iclr2024_conference,times}
\usepackage{todonotes}
\usepackage{graphicx}
\usepackage{subcaption}
\usepackage[inkscapelatex=false]{svg}

\usepackage{amsmath,amsfonts,bm}

\def\eqref#1{equation~\ref{#1}}

\def\1{\bm{1}}

\def\vx{{\bm{x}}}

\DeclareMathAlphabet{\mathsfit}{\encodingdefault}{\sfdefault}{m}{sl}
\SetMathAlphabet{\mathsfit}{bold}{\encodingdefault}{\sfdefault}{bx}{n}

\usepackage{hyperref}
\usepackage{url}

\title{Generating Interpretable Networks using \\ Hypernetworks}

\author{Isaac Liao \\
MIT \\
\texttt{iliao2345@gmail.com} \And
Ziming Liu \\
MIT \\
\texttt{zmliu@mit.edu} \And
Max Tegmark \\
MIT \\
\texttt{tegmark@mit.edu}
}

\iclrfinalcopy 
\begin{document}

\maketitle

\begin{abstract}

An essential goal in mechanistic interpretability to \textit{decode} a network, i.e., to convert a neural network's raw weights to an interpretable algorithm. Given the difficulty of the decoding problem, progress has been made to understand the easier \textit{encoding} problem, i.e., to convert an interpretable algorithm into network weights. Previous works focus on encoding existing algorithms into networks, which are interpretable by definition. However, focusing on encoding limits the possibility of discovering new algorithms that humans have never stumbled upon, but that are nevertheless interpretable. In this work, we explore the possibility of using hypernetworks to generate interpretable networks whose underlying algorithms are not yet known. The hypernetwork is carefully designed such that it can control network complexity, leading to a diverse family of interpretable algorithms ranked by their complexity. All of them are interpretable in hindsight, although some of them are less intuitive to humans, hence providing new insights regarding how to “think” like a neural network. For the task of computing L1 norms, hypernetworks find three algorithms: (a) the double-sided algorithm, (b) the convexity algorithm, (c) the pudding algorithm, although only the first algorithm was expected by the authors before experiments. We automatically classify these algorithms and analyze how these algorithmic phases develop during training, as well as how they are affected by complexity control. Furthermore, we show that a trained hypernetwork can correctly construct models for input dimensions not seen in training, demonstrating systematic generalization. 

\end{abstract}

\section{Introduction}

Although large language models have demonstrated a number of surprising mathematical and algorithmic capabilities \citep{yuan2023well,wei2022emergent}, it remains unknown whether they rediscover algorithms familiar to humans, or if they create more alien forms of mathematics and algorithms that appear less intuitive to humans. This question can be partially answered by recent efforts to  mechanistically interpret neural networks~\citep{scherlis2022polysemanticity,o2023disentangling,schubert2021high-low,power2022grokking,nanda2023progress,zhong2023clock}. The holy grail of mechanistic interpretability is to \textit{decode} model weights into interpretable algorithms. This is quite challenging because we have limited clues as to where to look and what to look for. Luckily, the inverse problem,  how to \textit{encode} an interpretable algorithm into model weights~\citep{lindner2023tracr}, may shed light on what an interpretable model may look like. Models converted from existing algorithms are by definition interpretable, but their limitations are also obvious: they rule out the possibility of new interpretable algorithms that no human has never stumbled upon (for whatever reason) but are nevertheless interpretable. 

This brings up a dilemma of mechanistic interpretability: a trained model is flexible but too uninterpretable, whereas a constructed model is interpretable but too inflexible. This raises the question of whether there is a way to balance between interpretability and flexibility. Ideally, we hope for models to reveal new algorithms that are undiscovered but which remain within the reach of human understanding. We propose to use hypernetworks~\citep{chauhan2023brief} for such a purpose. Intuitively, we find that hypernetworks are well-suited to this task because: (1) hypernetworks can generate “regular patterns of weights”, which are similar to the notion of interpretability; and (2) hypernetworks enable control over model complexity, so instead of generating one interpretable network, hypernetworks can generate a diverse family of networks with varying degrees of complexity.~\footnote{See Appendix~\ref{app:hypernetwork-intuition} for more discussion.}

\begin{figure}[tbp]
    \centering
    \includegraphics[width=0.9\textwidth]{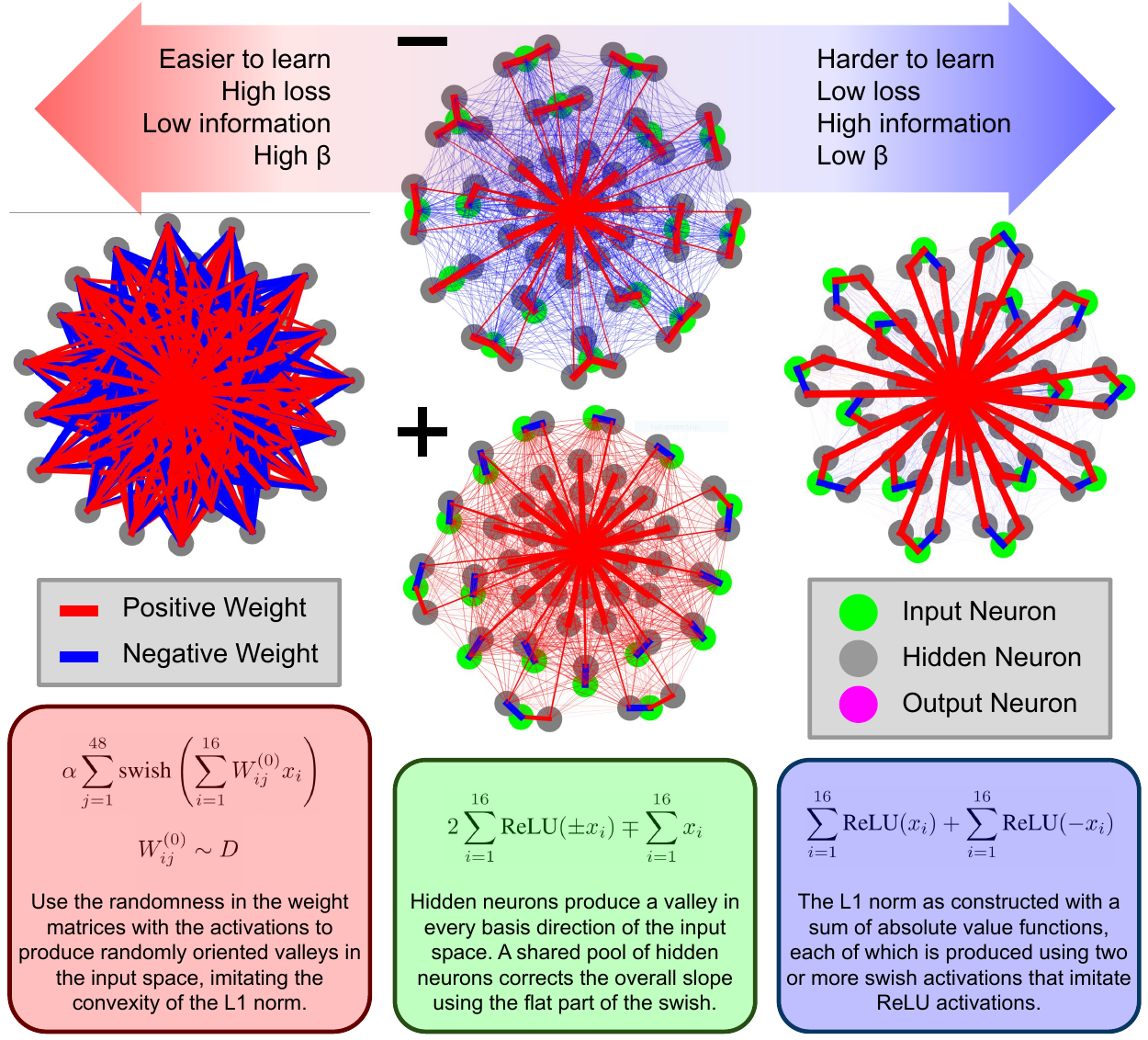}
    \caption{Three algorithms for computing the L1 norm, discovered by the hypernetwork. \textbf{Left: }the convexity algorithm. \textbf{Center top: }the negative pudding algorithm. \textbf{Center bottom: }the positive pudding algorithm. \textbf{Right: }the double-sided algorithm. The output neuron is in the center of all visualizations. The visualization method is included in Appendix~\ref{sec:graph_drawing}.}
    \label{fig:strategies}
    \vskip -0.2cm
\end{figure}

We focus on the simple example task of computing the $L_1$ norm of a vector. Although this task seems extremely simple and feels fully understood, our hypernetwork is able to generate new algorithms which appear less intuitive to humans yet still remain interpretable with a little overhead of reverse engineering. These new algorithms shed light on how neural networks may do computation or process information in ways that are different from humans. 
In particular, we identify in our neural networks three types of algorithms for computing the $L_1$ norm: the double-sided algorithm, the pudding algorithm, and the convexity algorithm (see Figure \ref{fig:strategies}), though the authors only expected the double-sided algorithm to be learned before experiments revealed otherwise. By contrast, a conventionally trained network appears to be highly uninterpretable, with no clear patterns in weights or activations (see Figure~\ref{fig:adam_comparison}). We further define order parameters to auto-classify these algorithms and find intriguing phase transitions between their occurrences, either in training, or when model complexity is varied. By ablating our hypernetwork, we also find that hypernetworks produce the pudding algorithm in two main ways, only one of which is disrupted by the ablation. We also show that a trained hypernetwork can correctly construct models for input dimensions not seen in training, demonstrating algorithmic generalization.  

Our results highlight the complexity of mechanistic descriptions even in models trained to perform extremely
simple mathematical tasks. We encourage the use of hypernetworks to help future works explore full algorithmic spaces in a controlled and systematic way.

\section{L1 Network Experiments}

In this section, we explain how all the networks we trained can compute the L1 norm. All of our networks are two-layer MLPs with 16 inputs, 48 hidden neurons, and 1 output, with swish activation \citep{ramachandran2017searching} \citep{hendrycks2016gaussian}. The training data is randomly generated online by sampling the inputs from i.i.d standard normal distribution, and by shifting and rescaling the target L1 norm outputs so that they have zero mean and unit variance.

We train a hypernetwork to generate good weights for this network. Our hypernetwork has a hyperparameter $\beta$ which controls the balance between the objectives of loss and model complexity (which we measure using a KL divergence). A higher $\beta$ values simplicity over reducing loss, and a lower $\beta$ values reducing loss over simplicity. Since our goal is to interpret the generated networks, we mostly treat the hypernetwork as a black box. Details for the internal structure of our hypernetwork are in Appendix \ref{sec:attentional_hypernetworks}. For our purposes, it suffices to remember that a hypernetwork can generate networks whose complexity can be controlled via $\beta$. As a result of this training, we have many models saved at various points in training for many $\beta$ values. Moreover, we tried 33 random seeds, so we have 33 independently sampled copies of all of these models. We find that three algorithms are discovered by the hypernetwork: the convexity algorithm, the pudding algorithm and the double-sided algorithm, shown in Figure~\ref{fig:strategies}.

\subsection{Interpretation of Generated Networks}

In this section, we deconstruct the algorithms performed by the networks generated by our hypernetworks. We determined these algorithms by looking at force-directed graph drawings of the learned networks \citep{kobourov2012spring}. Force-directed graph drawings are a way to visualize graphs of computations by organizing the nodes on the plane of a drawing, to make diagrams of these graphs more intuitive to read. Our force-directed graph drawings assign a position on a drawing to every neuron to minimize an energy consisting of mutual repulsion, connection strength weighted attraction, and central attraction \citep{bannister2013force}. Full details of how we generate these drawings can be found in Appendix \ref{sec:graph_drawing}. One can alternatively choose other visualization methods, but we find force-directly graph drawings especially useful since they make symmetries explicit (see Figure~\ref{fig:strategies}) hence one can simply distinguish different algorithms by noticing their symmetries and other visual traits. 
By looking at these drawings, we found that networks generally compute the L1 norm using one of three main algorithms:

\begin{itemize}
    \item \textbf{The Double-sided Algorithm} An absolute value function can be constructed with two ReLU neurons, i.e., $|x| = {\rm ReLU}(x) + {\rm ReLU}(-x)$. It is thus reasonable to expect a neural network to perform L1 computation by summing absolute values of all dimensions~\footnote{Note that we are actually using the SiLU activation, but SiLU and ReLU share similar qualitative behavior: a zoomed out plot of a swish function looks like a ReLU.}:
    \begin{equation}
    ||\vx||_1 = \sum_{i=1}^{16} \text{ReLU}(x_i) + \sum_{i=1}^{16} \text{ReLU}(-x_i)
    \end{equation}
    Indeed, this is one possible algorithm that the hypernetwork produces.

    \item \textbf{The (Signed) Pudding Algorithm} It turns out that there is another method of computing the L1 norm using ReLUs. The hypernetworks that generate the pudding algorithm have learned to take advantage of the following fact:
    \begin{align}
    ||\vx||_1 =& 2\sum_{i=1}^{16} \text{ReLU}(\mp x_i) \pm \sum_{i=1}^{16} x_i \\
    \approx& \lim_{c\to \infty} 2\sum_{j=1}^{16} \text{ReLU}(\mp x_j \pm \sum_{i=1}^{16} x_i) + \sum_{j=1}^{32} \left(\text{ReLU}(c \pm \sum_{i=1}^{16}x_i)-c\right) \label{eq:imperfect_pudding}
    \end{align}
    which holds for both signs $\pm$ (hence the name “signed pudding”), where $i$ iterates through input neurons and $j$ through hidden neurons. The signed pudding algorithm assigns one hidden neuron to each input neuron $i$ to compute the first summation, and uses the leftover hidden neurons to compute the second summation term. Note that only one neuron is actually needed to compute the second term, and so this algorithm can actually be implemented with $n+1$ hidden neurons, which is more efficient than the $2n$ needed for the double-sided algorithm. The pudding algorithm is almost always implemented imperfectly, as in Equation \ref{eq:imperfect_pudding}. The hypernetwork uses a hard-coded assignment of pairs of hidden neurons to input neurons; changing the generation seed does not change the order of the hidden neurons. This is the most common algorithm found in our experiments.

    \item \textbf{The Convexity Algorithm} This is an imperfect random algorithm that is easy for the hypernetwork to produce. This algorithm notices that the L1 norm is a convex function, and it tries to match the convexity using randomly oriented swish functions:
    \begin{equation}
    ||\vx||_1 \approx \alpha\sum_{j=1}^{48} \text{swish}\left(\sum_{i=1}^{16}W_{ij}^{(0)}x_i\right), \quad W_{ij}^{(0)} \sim D
    \end{equation}
    with $\alpha$ some constant and $D$ some distribution that is typically symmetric. Sometimes the distribution of $D$ is unimodal, and sometimes it is bimodal.
\end{itemize}

\subsection{Order Parameters}
Having identified that our experiments mainly consist of three algorithms, we construct a number of order parameters below to distinguish the three algorithms apart from one another.

\textbf{Double Sidedness:} Given the weight matrix $W \in \mathbb{R}^{n_0 \times n_1}$ for the first linear layer (bias not included) where $n_0$ is the number of input neurons and $n_1$ is the number of hidden neurons, the double sidedness order parameter $\alpha_1$ is defined as:
\begin{equation}
\alpha_1 = \frac{\underset{i}{\min\,} \min(-\underset{j}{\min\,}W_{ij}, \underset{j}{\max\,}W_{ij})}{\underset{i,j}{\text{median}\,}\text{abs}(W_{ij})}
\end{equation}
The double sidedness measures the degree to which the solution uses the double-sided algorithm.

\textbf{Strongest Connection:} The strongest connection order parameter $\alpha_2$ is defined as:
\begin{equation}
\alpha_2 = \underset{i,j}{\max\,}\text{abs}(W_{ij})
\end{equation}
A weak strongest connection is indicative of the convexity solution.

\begin{figure}
    \centering
    \includegraphics[width=\textwidth]{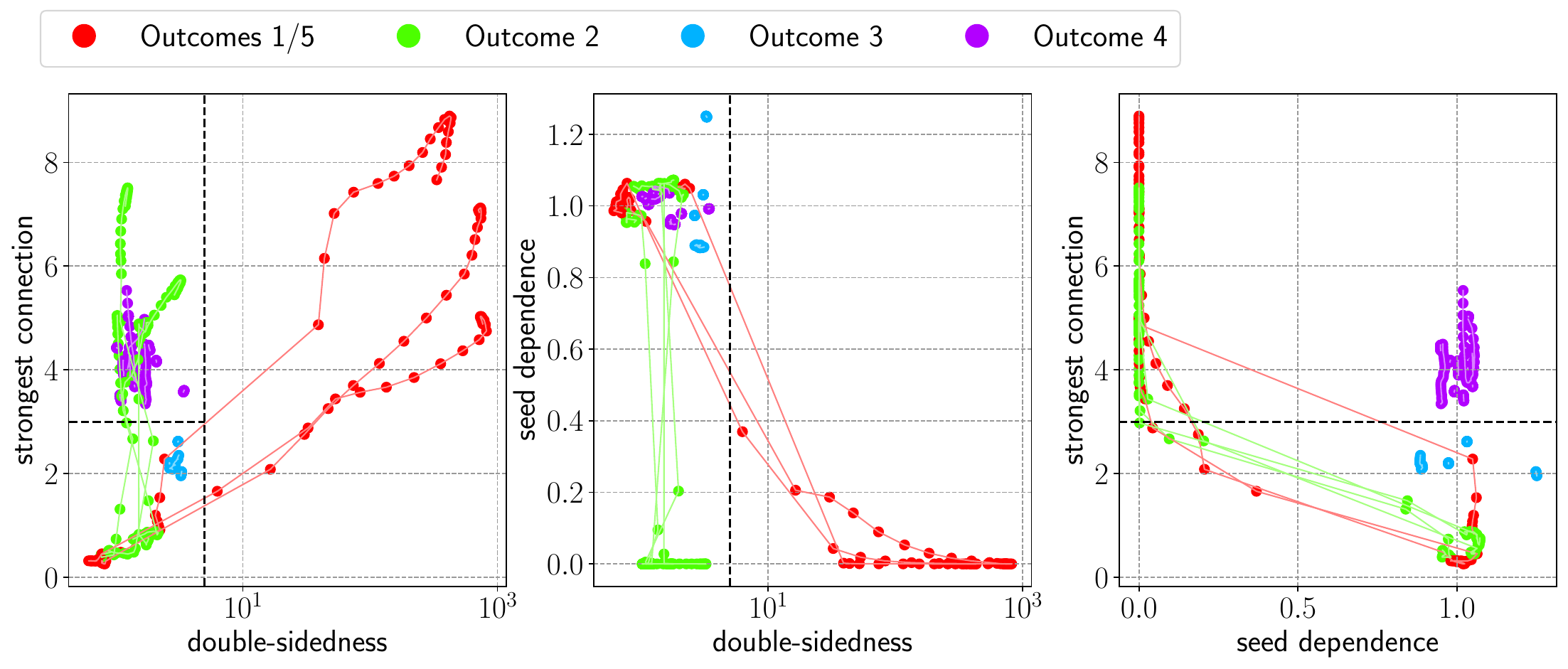}
    \caption{Networks that try to compute the L1 norm cluster into three general algorithms in order parameter space, spanned by the “strongest connection”, “double-sidedness” and “seed dependence” order parameters. Lines are generated by sweeping $\beta$ from $10^{-12}$ to $1$ in 30 increments logarithmically. The dotted lines represent hand-picked boundaries which determine when phase transitions between algorithms occur. Lines are grouped and colored by the phases where they start and end at.}
    \label{fig:order_parameters}
\end{figure}  

\textbf{Seed Dependence:} The seed dependence order parameter $\alpha_3$ is defined as:
\begin{equation}
\frac{||W-V||_F^2}{||W||_F^2+||V||_F^2}
\end{equation}
where $W$ and $V$ are weight matrices for the first linear layer (bias not included) generated with different randomization seeds using the same hypernetwork. A low seed dependence indicates that the hypernetwork is using memorized information about configurations of weights rather than generating this information randomly.

The networks divide themselves roughly into three clusters in order parameter space, each corresponding to one algorithm, as shown in Figure \ref{fig:order_parameters}. While the seed dependence is not immediately relevant since it does not differentiate between algorithms, we will explain later that it can be used to investigate the way that the hypernetwork constructs the pudding algorithm.



\subsection{Development of Algorithms Throughout Training}

\begin{figure}[ht]
    \centering
    \begin{subfigure}{0.32\textwidth}
        \centering
        \includegraphics[width=\linewidth]{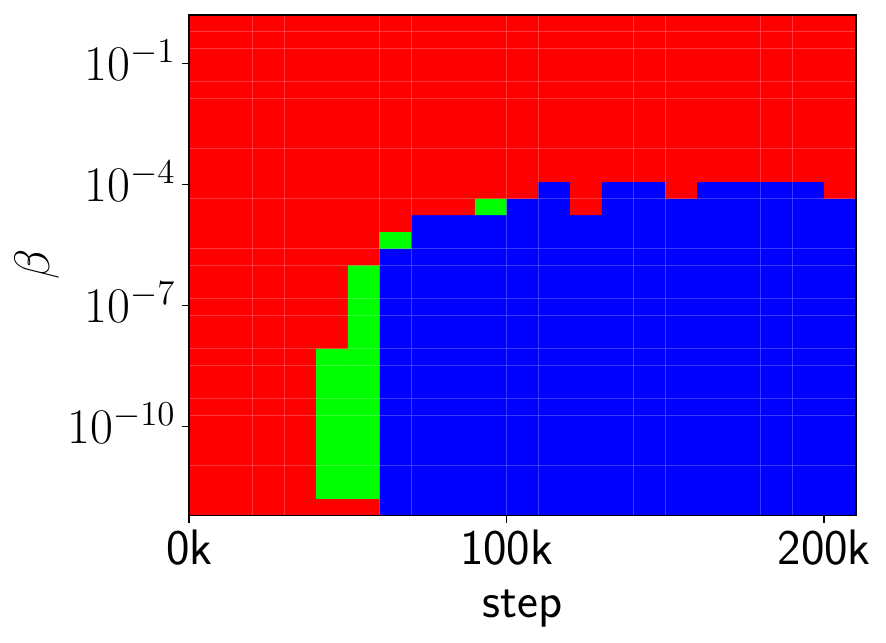}
        \caption{Outcome 1}
        \label{subfig:strategy_1}
    \end{subfigure}
    \begin{subfigure}{0.32\textwidth}
        \centering
        \includegraphics[width=\linewidth]{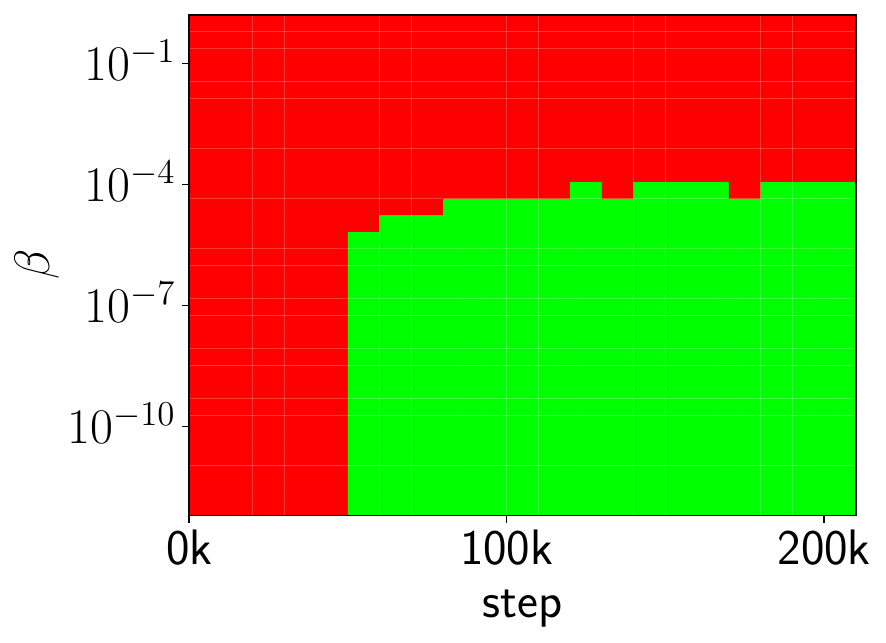}
        \caption{Outcome 2}
        \label{subfig:strategy_2}
    \end{subfigure}
    \begin{subfigure}{0.32\textwidth}
        \centering
        \includegraphics[width=\linewidth]{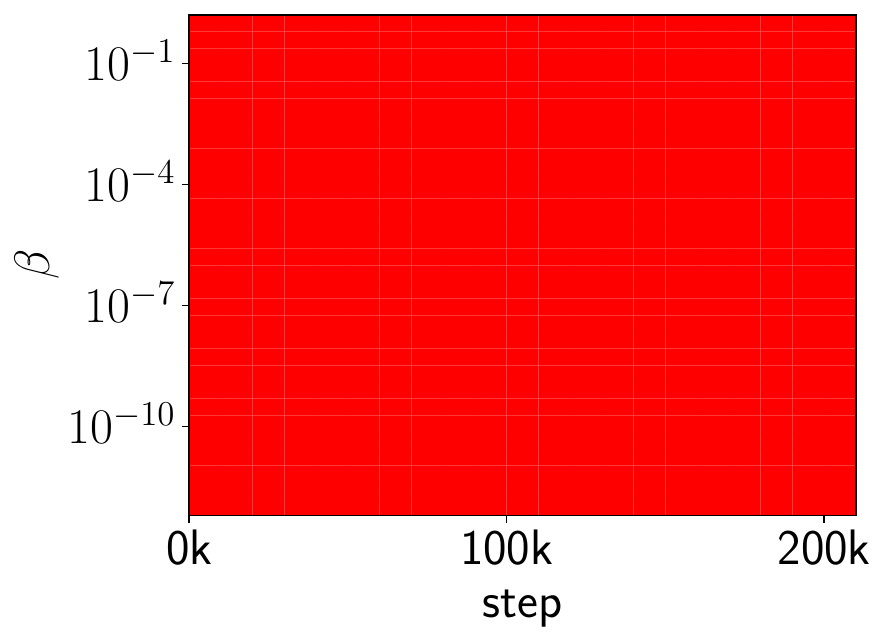}
        \caption{Outcome 3}
        \label{subfig:strategy_3}
    \end{subfigure}
    
    \begin{subfigure}{0.32\textwidth}
        \centering
        \includegraphics[width=\linewidth]{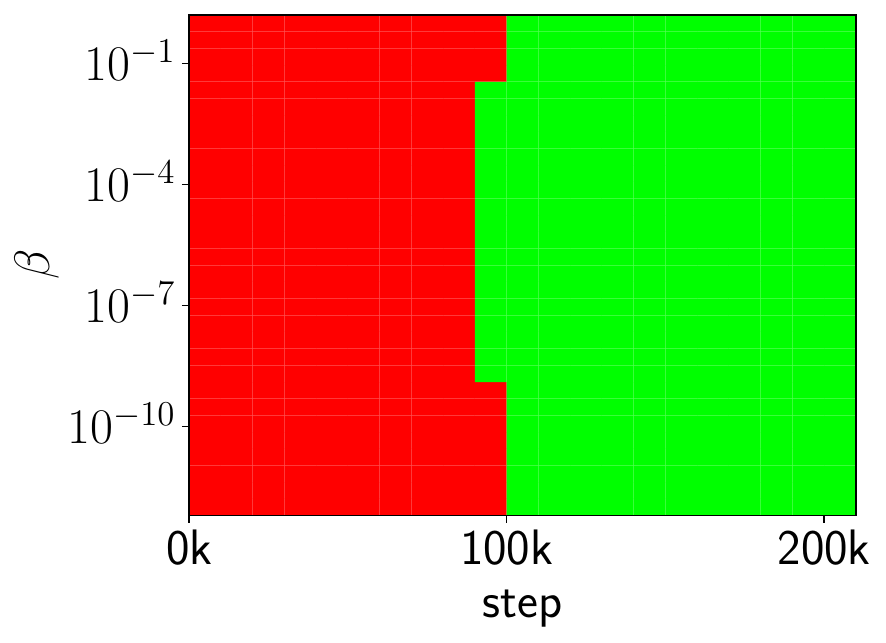}
        \caption{Outcome 4}
        \label{subfig:strategy_4}
    \end{subfigure}
    \begin{subfigure}{0.32\textwidth}
        \centering
        \includegraphics[width=\linewidth]{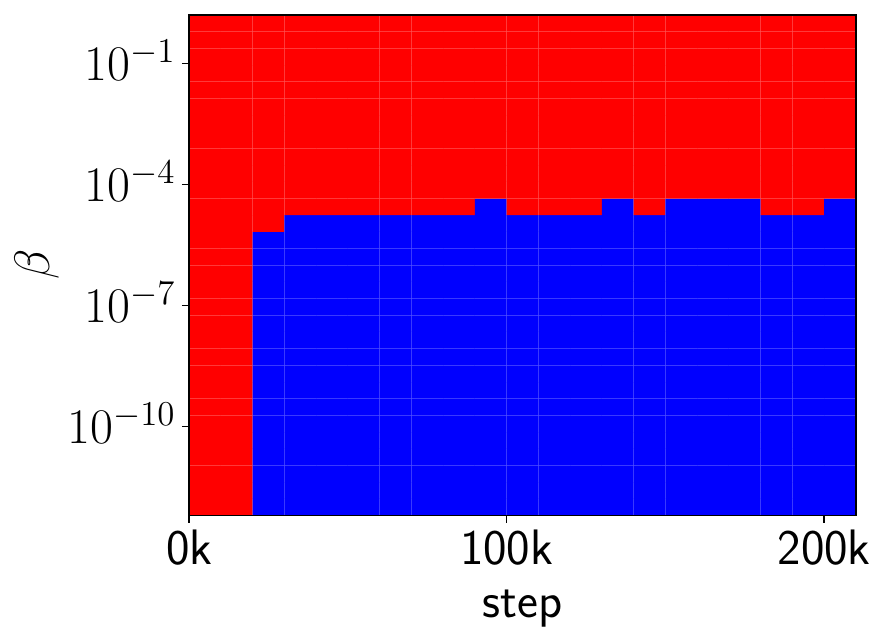}
        \caption{Outcome 5}
        \label{subfig:strategy_5}
    \end{subfigure}
    
    \caption{Five ways to develop the three algorithms through training. The horizontal axis is the step number, while the vertical axis is the $\beta$ parameter used to generate the network. \textbf{Red:} convexity algorithm. \textbf{Green:} pudding algorithm. \textbf{Blue:} double-sided algorithm.}
    \label{fig:strategy_development}
\end{figure}

Using the order parameters, we automatically classified the algorithms which developed at various points during training for various $\beta$ values, as in Figure \ref{fig:strategy_development}. We find that the convexity algorithm always develops first, and that other algorithms differentiate away from there. The convexity algorithm can evolve into either the pudding or double-sided algorithms, and the pudding algorithm sometimes also transitions into the double-sided algorithm. Transitions to the pudding algorithm can happen either for only low $\beta$ or for all $\beta$. Oftentimes, the high $\beta$ regime retains the convexity algorithm while the low $\beta$ regime evolves though multiple algorithms, and the $\beta$ value of the transition boundary increases over time and then stabilizes.



\begin{figure}
    \centering
    \includegraphics[width=0.7\textwidth]{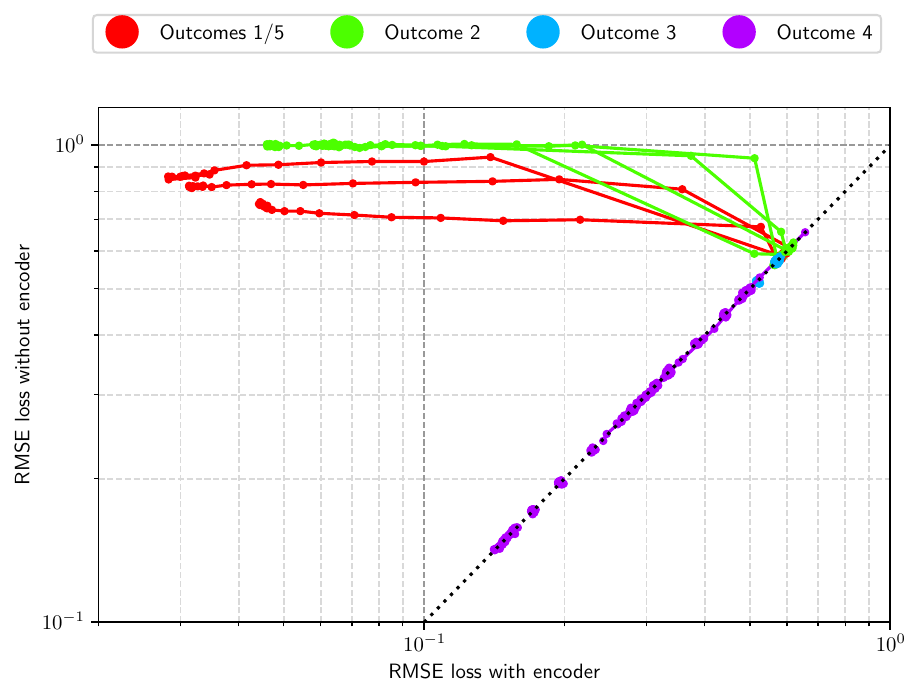}
    \caption{Loss of neural network generated when the encoder is either used or ignored. The black dotted line indicates when the performance is unaffected by the presence of the encoder, i.e., when the encoder is unused during the weight generation process. Lines are generated by sweeping $\beta$ from $10^{-12}$ to $1$ in 30 logarithmic increments.}
    \label{fig:with_vs_without_encoder}
\end{figure}

It is worth noting that in Figure (\ref{subfig:strategy_4}) the hypernetwork becomes insensitive to $\beta$. In this case, the hypernetwork's accumulated KL divergence sums up to nearly zero for all $\beta$, causing the Pareto frontier between loss and simplicity to collapse to a single point. This is not the case for the other cases that develop the pudding algorithm; the KL divergence usually increases considerably for lower $\beta$. We believe that in case (\ref{subfig:strategy_4}), the hypernetwork randomly generates an assignment of hidden neurons to input neurons, while in the other cases, the hypernetwork stores a memorized assignment on the decoder side and passes it through to the encoder for output. Thus, the hypernetwork accumulates KL in all cases except (\ref{subfig:strategy_4}). To test this hypothesis, we generated another network using only the decoder side of the hypernetwork without the encoder side\footnote{by drawing latents from the distribution defined by the decoder side instead of the encoder side}, and evaluated its loss on the L1 problem again. Removing the encoder should prevent the hypernetwork from using any memorized assignments, without affecting its ability to randomly generate an assignment. Indeed, Figure \ref{fig:with_vs_without_encoder} shows that when we remove the encoder, hypernetworks in case (\ref{subfig:strategy_4}) are still able to construct working L1 networks that implement the pudding algorithm, but hypernetworks in other cases are not.


\subsection{Generalization Capabilities}
\begin{figure}
    \centering
    \includegraphics[width=0.7\textwidth]{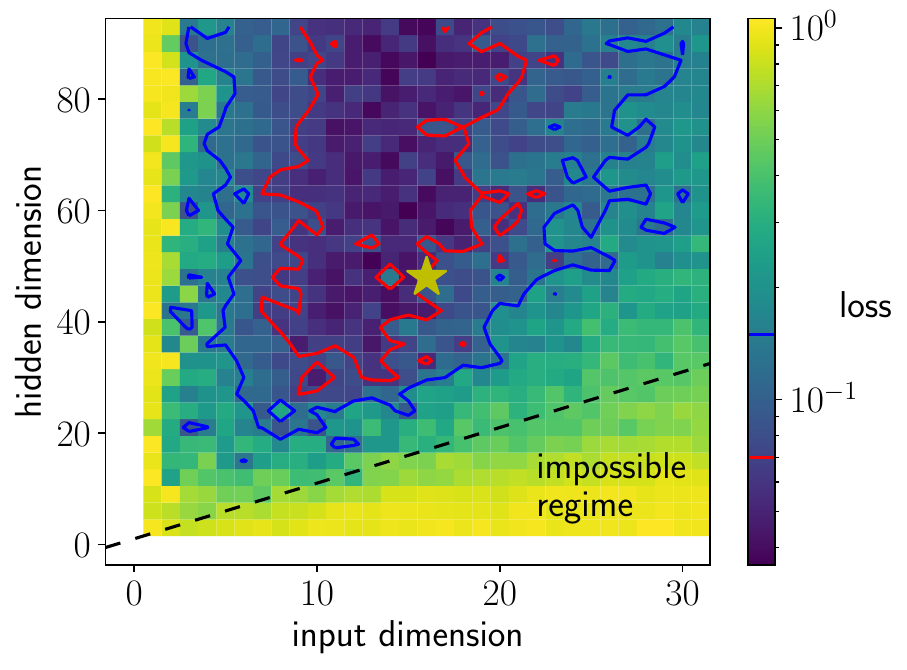}
    \caption{Loss of neural network generated with various input dimensions and hidden dimensions using a hypernetwork. The red contour denotes a loss of $0.07$, while the blue denotes a loss of $0.15$. The hypernetwork was only trained to generate networks with input dimension 16 and hidden dimension 48 (yellow star), yet it can produce networks of diverse shapes which all compute the L1 norm with reasonable accuracy.}
    \label{fig:generalization}
\end{figure}

A key feature of the assignment generation hypothesis is that implies hypernetworks in case (\ref{subfig:strategy_4}) can generate L1 networks of different layer sizes, because all the configurations of weights are automatically randomly generated instead of being memorized specifically for the $(16, 48, 1)$ layer size structure. We can therefore use an existing hypernetwork to generate networks that compute the L1 norm in a wide range of dimensions, including dimensions larger than 16 which is what the hypernetwork was trained for. The hidden layer size can also be modified to be larger or smaller in the same way. Figure \ref{fig:generalization} shows that many of these L1 networks perform similarly to the original $(16, 48, 1)$ network. We find that there is a region of low loss which extends in the direction of increasing input dimension and hidden dimension, leading us to believe that the hypernetwork has found a general algorithm for computing L1 norms of vectors of any arbitrarily large size, demonstrating systematic generalization.

\subsection{Baseline Algorithm}
\begin{figure}[h!]
    \centering
    \begin{subfigure}{0.45\textwidth}
        \centering
        \includegraphics[width=\linewidth]{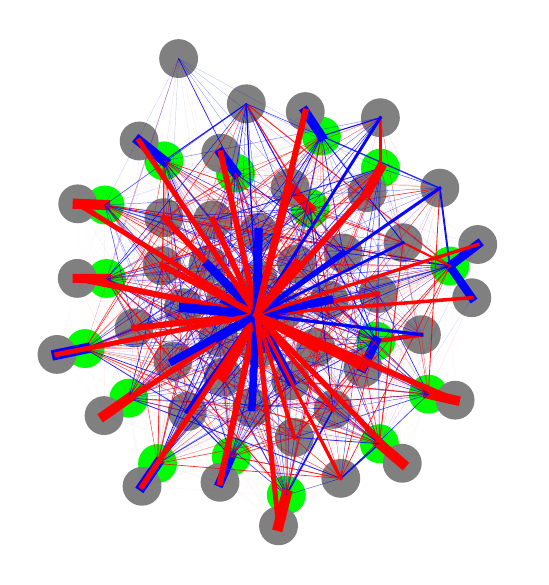}
        \caption{Trained by Adam.}
        \label{subfig:strategy_1}
    \end{subfigure}
    \hfill
    \begin{subfigure}{0.45\textwidth}
        \centering
        \includegraphics[width=\linewidth]{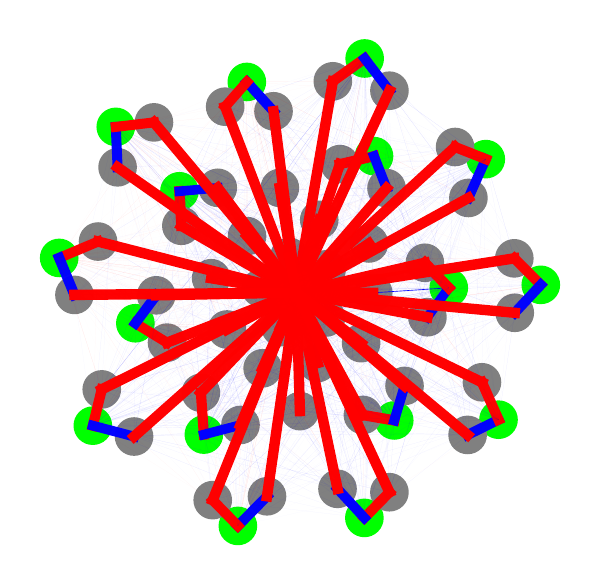}
        \caption{Generated via hypernetwork.}
        \label{subfig:strategy_2}
    \end{subfigure}
    \caption{Visualization of a $(16, 48, 1)$ neural networks trained to compute the L1 norm of a vector. Input neurons in green, output neurons in magenta. Positive weights in red, and negative weights in blue. The Adam-trained network is messy, whereas the hypernetwork-generated network is much easier to interpret.}
    \label{fig:adam_comparison}
\end{figure}

As a baseline, we used Adam \citep{kingma2014adam} to train the same L1 norm network instead of generating the weights using a hypernetwork. The resulting network has a lower loss than the networks generated via hypernetwork, but it is much more difficult to interpret. This is already visible in the cleanliness of the visualization for the hypernetwork in comparison to Adam, as shown in Figure \ref{fig:adam_comparison}.

Since the Adam-trained L1 norm network is much harder to interpret, we only have a hypothesis about how it computes the L1 norm, which is as follows. In the Adam-trained L1 norm network, hidden neurons are split into two groups: some that are assigned to a single input neuron, and the leftover hidden neurons that are attached to all input neurons. Each input neuron then uses any of its assigned hidden neuron(s) to create a kink of appropriate difference in slope between the positive and negative sides, and the leftover hidden neurons are used to correct the average slope between the two sides, forming an absolute value function for that input neuron. The absolute value functions for each input neuron are then added together. This is like the pudding algorithm except that there can be many hidden neurons assigned to one input neuron (e.g., 5 hidden neurons, $+++--$) and that the corresponding weights can differ in their sign. Signs for all weights connecting to any given leftover hidden neuron are randomized, and each input neuron connects to every leftover hidden neuron with a different random strength, provided that all the weights satisfy the constraints above. The takeaway is that the network trained by Adam is unnecessarily complex and variable due to the randomness involved, whereas a much simpler and more interpretable solution exists that can be found via hypernetworks.

\section{Related Work}


{\bf Hypernetworks} Most current hypernetwork research focuses on how we can use hypernetworks to predict different attributes of a model in a particular setting without having to train it in that setting, for example it's loss, accuracy, or trained parameters \citep{zhang2018graph} \citep{knyazev2021parameter}. This information can be used to design an architecture with lower loss, skip some training steps \citep{knyazev2023can}, adjust the parameters based on different loss functions \citep{navon2020learning}, and more. Our work instead uses hypernetworks to compress and generate data that comes in the form of neural network weights. Training hypernetworks typically involves computing hypergradients \citep{baydin2017online}, and our work is no exception to this.

{\bf Minimum description length} The minimum description length (MDL) principle is a mathematical version of Occam's razor that prefers the most compressed explanation for a given dataset \citep{mdlbook,grunwald2019minimum,RISSANEN1978465,Solomonoff2009}. The MDL principle can treat a neural network's learning as the process of information compression, so that the complexity of a model is expressed by various KL divergences that form the loss to be minimized \citep{10.1145/321892.321894,polyanskiytextbook}. The MDL principle implies that our hypernetwork is trying to find weights which are as simple as possible for solving the task. The weights' simplicity makes them easier for humans to pick apart: the resulting model should be very interpretable, and this fact is the main driving force of this paper.


{\bf Neural network compression} While we were building a compressor to compress neural networks, we noticed that most machine learning systems compress sets of weights with techniques \citep{51791361-8fe2-38d5-959f-ae8d048b490d,frankle2018lottery,tan2019efficientnet,white2023neural,redmon2018yolov3} that are different to how data from a dataset is typically compressed \citep{kingma2013auto,kobyzev2020normalizing,vahdat2020nvae,sonderby2016ladder,child2020very,vaswani2017attention,openai2023gpt4,touvron2023llama}. When we tried to use techniques for data compression to compress weights instead, it turned out we were constructing systems that are commonly known as hypernetworks \citep{chauhan2023brief}. Our hypernetwork's architecture is based on graph neural networks, self-attention, and deep hierarchical VAEs, all in combination.


{\bf Mechanistic Interpretability} Research on mechanistic interpretability helps us explain how neural networks operate at the individual neuron level, so that we can understand why they produce certain outputs. This lets us build safer models that we can have better trust in for applications that need this trust. Landmark works in interpretability have included the discovery of polysemantic neurons \citep{scherlis2022polysemanticity,o2023disentangling}, high-low frequency detectors \citep{schubert2021high-low}, edge detectors, arithmetic representations~\citep{power2022grokking}, modularity~\citep{liu2023seeing}, and algorithmic circuits~\citep{nanda2023progress,zhong2023clock,wang2022interpretability}.





\section{Conclusion and Discussion}


In this paper, we have introduced a novel hypernetwork-based method for constructing neural network which makes them mechanistically interpretable. We then used this method to construct networks which compute the L1 norm and mechanistically interpreted them.

We found that the hypernetwork-generated L1 norm networks implement three main algorithms for computing the L1 norm, and they represent different tradeoffs between their errors and the model simplicity. This tradeoff can be manipulated by the $\beta$ hyperparameter, which controls the relative weight of error vs. simplicity. The algorithm most expected by humans is the double-sided algorithm, which is the hardest to learn and the most accurate. We constructed three order parameters, two of which we use to automatically classify neural networks according to the three algorithms. We find that the three algorithms develop in different $\beta$ regimes and at different times during training. Namely, the convexity algorithm is the easiest to learn, simplest, and the one that develops at greatest $\beta$, followed by the pudding algorithm, followed by the double-sided algorithm. The pudding algorithm is even more simple than the expected double-sided, in that it can be used to compute the L1 norm with fewer hidden neurons, than what the authors originally thought was possible.

There is also value in developing explanations for how the hypernetworks themselves learn to build neural networks. We find that the hypernetworks develop two main methods for constructing networks which operate via the pudding algorithm: one which constructs a random assignment of hidden neurons to input neurons and another which uses a memorized assignment whose information content is penalized. We demonstrate that hypernetworks which construct random assignments can be used to generate working L1 networks to operate on different, sometimes even larger input sizes and hidden dimensions. This works completely in inference time, without any retraining. The hypernetwork's ability to generalize to other input dimensions signals that the hypernetwork has learned an algorithm for computing L1 norms in general for any input vector size, rather than just a circuit that computes an L1 norm with a fixed input size that cannot generalize to other sizes. This work is a preliminary report that showcases the potential of hypernetworks for interpretability research. In the future we hope to generalize the analysis of hypernetwork to more complicated realistic problems.


\section{Acknowledgements}
We would like to acknowledge the MIT SuperCloud and Lincoln Laboratory Supercomputing Center for providing HPC resources that have contributed to the research results reported within this paper. This work was also sponsored in part by the National Science Foundation under Cooperative Agreement PHY-2019786 (The NSF AI Institute for Artificial Intelligence and Fundamental Interactions, \url{http://iaifi.org/}) as well as the Beneficial AI Foundation.

\bibliography{iclr2024_conference}

\begin{thebibliography}{48}
\providecommand{\natexlab}[1]{#1}
\providecommand{\url}[1]{\texttt{#1}}
\expandafter\ifx\csname urlstyle\endcsname\relax
  \providecommand{\doi}[1]{doi: #1}\else
  \providecommand{\doi}{doi: \begingroup \urlstyle{rm}\Url}\fi

\bibitem[Bannister et~al.(2013)Bannister, Eppstein, Goodrich, and Trott]{bannister2013force}
Michael~J Bannister, David Eppstein, Michael~T Goodrich, and Lowell Trott.
\newblock Force-directed graph drawing using social gravity and scaling.
\newblock In \emph{Graph Drawing: 20th International Symposium, GD 2012, Redmond, WA, USA, September 19-21, 2012, Revised Selected Papers 20}, pp.\  414--425. Springer, 2013.

\bibitem[Baydin et~al.(2017)Baydin, Cornish, Rubio, Schmidt, and Wood]{baydin2017online}
Atilim~Gunes Baydin, Robert Cornish, David~Martinez Rubio, Mark Schmidt, and Frank Wood.
\newblock Online learning rate adaptation with hypergradient descent.
\newblock \emph{arXiv preprint arXiv:1703.04782}, 2017.

\bibitem[Chaitin(1975)]{10.1145/321892.321894}
Gregory~J. Chaitin.
\newblock A theory of program size formally identical to information theory.
\newblock \emph{J. ACM}, 22\penalty0 (3):\penalty0 329–340, jul 1975.
\newblock ISSN 0004-5411.
\newblock \doi{10.1145/321892.321894}.
\newblock URL \url{https://doi.org/10.1145/321892.321894}.

\bibitem[Chauhan et~al.(2023)Chauhan, Zhou, Lu, Molaei, and Clifton]{chauhan2023brief}
Vinod~Kumar Chauhan, Jiandong Zhou, Ping Lu, Soheila Molaei, and David~A Clifton.
\newblock A brief review of hypernetworks in deep learning.
\newblock \emph{arXiv preprint arXiv:2306.06955}, 2023.

\bibitem[Child(2020)]{child2020very}
Rewon Child.
\newblock Very deep vaes generalize autoregressive models and can outperform them on images.
\newblock \emph{arXiv preprint arXiv:2011.10650}, 2020.

\bibitem[Daigavane et~al.(2021)Daigavane, Ravindran, and Aggarwal]{daigavane2021understanding}
Ameya Daigavane, Balaraman Ravindran, and Gaurav Aggarwal.
\newblock Understanding convolutions on graphs.
\newblock \emph{Distill}, 2021.
\newblock \doi{10.23915/distill.00032}.
\newblock https://distill.pub/2021/understanding-gnns.

\bibitem[Frankle \& Carbin(2018)Frankle and Carbin]{frankle2018lottery}
Jonathan Frankle and Michael Carbin.
\newblock The lottery ticket hypothesis: Finding sparse, trainable neural networks.
\newblock \emph{arXiv preprint arXiv:1803.03635}, 2018.

\bibitem[Gr{\"u}nwald \& Roos(2019)Gr{\"u}nwald and Roos]{grunwald2019minimum}
Peter Gr{\"u}nwald and Teemu Roos.
\newblock Minimum description length revisited.
\newblock \emph{International journal of mathematics for industry}, 11\penalty0 (01):\penalty0 1930001, 2019.

\bibitem[Grünwald(2007)]{mdlbook}
Peter Grünwald.
\newblock \emph{The Minimum Description Length Principle}.
\newblock 01 2007.
\newblock ISBN 9780262256292.
\newblock \doi{10.7551/mitpress/4643.001.0001}.

\bibitem[Hendrycks \& Gimpel(2016)Hendrycks and Gimpel]{hendrycks2016gaussian}
Dan Hendrycks and Kevin Gimpel.
\newblock Gaussian error linear units (gelus).
\newblock \emph{arXiv preprint arXiv:1606.08415}, 2016.

\bibitem[Higgins et~al.(2017)Higgins, Matthey, Pal, Burgess, Glorot, Botvinick, Mohamed, and Lerchner]{higgins2017betavae}
Irina Higgins, Loic Matthey, Arka Pal, Christopher Burgess, Xavier Glorot, Matthew Botvinick, Shakir Mohamed, and Alexander Lerchner.
\newblock beta-{VAE}: Learning basic visual concepts with a constrained variational framework.
\newblock In \emph{International Conference on Learning Representations}, 2017.
\newblock URL \url{https://openreview.net/forum?id=Sy2fzU9gl}.

\bibitem[Ho et~al.(2020)Ho, Jain, and Abbeel]{ho2020denoising}
Jonathan Ho, Ajay Jain, and Pieter Abbeel.
\newblock Denoising diffusion probabilistic models.
\newblock \emph{Advances in neural information processing systems}, 33:\penalty0 6840--6851, 2020.

\bibitem[Kingma \& Ba(2014)Kingma and Ba]{kingma2014adam}
Diederik~P Kingma and Jimmy Ba.
\newblock Adam: A method for stochastic optimization.
\newblock \emph{arXiv preprint arXiv:1412.6980}, 2014.

\bibitem[Kingma \& Welling(2013)Kingma and Welling]{kingma2013auto}
Diederik~P Kingma and Max Welling.
\newblock Auto-encoding variational bayes.
\newblock \emph{arXiv preprint arXiv:1312.6114}, 2013.

\bibitem[Knyazev et~al.(2021)Knyazev, Drozdzal, Taylor, and Romero~Soriano]{knyazev2021parameter}
Boris Knyazev, Michal Drozdzal, Graham~W Taylor, and Adriana Romero~Soriano.
\newblock Parameter prediction for unseen deep architectures.
\newblock \emph{Advances in Neural Information Processing Systems}, 34:\penalty0 29433--29448, 2021.

\bibitem[Knyazev et~al.(2023)Knyazev, Hwang, and Lacoste-Julien]{knyazev2023can}
Boris Knyazev, Doha Hwang, and Simon Lacoste-Julien.
\newblock Can we scale transformers to predict parameters of diverse imagenet models?
\newblock \emph{arXiv preprint arXiv:2303.04143}, 2023.

\bibitem[Kobourov(2012)]{kobourov2012spring}
Stephen~G Kobourov.
\newblock Spring embedders and force directed graph drawing algorithms.
\newblock \emph{arXiv preprint arXiv:1201.3011}, 2012.

\bibitem[Kobyzev et~al.(2020)Kobyzev, Prince, and Brubaker]{kobyzev2020normalizing}
Ivan Kobyzev, Simon~JD Prince, and Marcus~A Brubaker.
\newblock Normalizing flows: An introduction and review of current methods.
\newblock \emph{IEEE transactions on pattern analysis and machine intelligence}, 43\penalty0 (11):\penalty0 3964--3979, 2020.

\bibitem[Lindner et~al.(2023)Lindner, Kram{\'a}r, Rahtz, McGrath, and Mikulik]{lindner2023tracr}
David Lindner, J{\'a}nos Kram{\'a}r, Matthew Rahtz, Thomas McGrath, and Vladimir Mikulik.
\newblock Tracr: Compiled transformers as a laboratory for interpretability.
\newblock \emph{arXiv preprint arXiv:2301.05062}, 2023.

\bibitem[Liu et~al.(2023)Liu, Gan, and Tegmark]{liu2023seeing}
Ziming Liu, Eric Gan, and Max Tegmark.
\newblock Seeing is believing: Brain-inspired modular training for mechanistic interpretability.
\newblock \emph{arXiv preprint arXiv:2305.08746}, 2023.

\bibitem[Nanda et~al.(2023)Nanda, Chan, Lieberum, Smith, and Steinhardt]{nanda2023progress}
Neel Nanda, Lawrence Chan, Tom Lieberum, Jess Smith, and Jacob Steinhardt.
\newblock Progress measures for grokking via mechanistic interpretability.
\newblock In \emph{The Eleventh International Conference on Learning Representations}, 2023.
\newblock URL \url{https://openreview.net/forum?id=9XFSbDPmdW}.

\bibitem[Navon et~al.(2020)Navon, Shamsian, Chechik, and Fetaya]{navon2020learning}
Aviv Navon, Aviv Shamsian, Gal Chechik, and Ethan Fetaya.
\newblock Learning the pareto front with hypernetworks.
\newblock \emph{arXiv preprint arXiv:2010.04104}, 2020.

\bibitem[O'Mahony et~al.(2023)O'Mahony, Andrearczyk, M{\"u}ller, and Graziani]{o2023disentangling}
Laura O'Mahony, Vincent Andrearczyk, Henning M{\"u}ller, and Mara Graziani.
\newblock Disentangling neuron representations with concept vectors.
\newblock In \emph{Proceedings of the IEEE/CVF Conference on Computer Vision and Pattern Recognition}, pp.\  3769--3774, 2023.

\bibitem[OpenAI(2023)]{openai2023gpt4}
OpenAI.
\newblock Gpt-4 technical report, 2023.

\bibitem[Polyanskiy \& Wu()Polyanskiy and Wu]{polyanskiytextbook}
Yury Polyanskiy and Yihong Wu.
\newblock Information theory: From coding to learning.
\newblock preprint on webpage at \url{https://people.lids.mit.edu/yp/homepage/papers.html}.

\bibitem[Power et~al.(2022)Power, Burda, Edwards, Babuschkin, and Misra]{power2022grokking}
Alethea Power, Yuri Burda, Harri Edwards, Igor Babuschkin, and Vedant Misra.
\newblock Grokking: Generalization beyond overfitting on small algorithmic datasets.
\newblock \emph{arXiv preprint arXiv:2201.02177}, 2022.

\bibitem[Ramachandran et~al.(2017)Ramachandran, Zoph, and Le]{ramachandran2017searching}
Prajit Ramachandran, Barret Zoph, and Quoc~V Le.
\newblock Searching for activation functions.
\newblock \emph{arXiv preprint arXiv:1710.05941}, 2017.

\bibitem[Redmon \& Farhadi(2018)Redmon and Farhadi]{redmon2018yolov3}
Joseph Redmon and Ali Farhadi.
\newblock Yolov3: An incremental improvement.
\newblock \emph{arXiv preprint arXiv:1804.02767}, 2018.

\bibitem[Rissanen(1978)]{RISSANEN1978465}
J.~Rissanen.
\newblock Modeling by shortest data description.
\newblock \emph{Automatica}, 14\penalty0 (5):\penalty0 465--471, 1978.
\newblock ISSN 0005-1098.
\newblock \doi{https://doi.org/10.1016/0005-1098(78)90005-5}.
\newblock URL \url{https://www.sciencedirect.com/science/article/pii/0005109878900055}.

\bibitem[Sanchez-Lengeling et~al.(2021)Sanchez-Lengeling, Reif, Pearce, and Wiltschko]{sanchez-lengeling2021a}
Benjamin Sanchez-Lengeling, Emily Reif, Adam Pearce, and Alexander~B. Wiltschko.
\newblock A gentle introduction to graph neural networks.
\newblock \emph{Distill}, 2021.
\newblock \doi{10.23915/distill.00033}.
\newblock https://distill.pub/2021/gnn-intro.

\bibitem[Scarselli et~al.(2009)Scarselli, Gori, Tsoi, Hagenbuchner, and Monfardini]{4700287}
Franco Scarselli, Marco Gori, Ah~Chung Tsoi, Markus Hagenbuchner, and Gabriele Monfardini.
\newblock The graph neural network model.
\newblock \emph{IEEE Transactions on Neural Networks}, 20\penalty0 (1):\penalty0 61--80, 2009.
\newblock \doi{10.1109/TNN.2008.2005605}.

\bibitem[Scherlis et~al.(2022)Scherlis, Sachan, Jermyn, Benton, and Shlegeris]{scherlis2022polysemanticity}
Adam Scherlis, Kshitij Sachan, Adam~S Jermyn, Joe Benton, and Buck Shlegeris.
\newblock Polysemanticity and capacity in neural networks.
\newblock \emph{arXiv preprint arXiv:2210.01892}, 2022.

\bibitem[Schubert et~al.(2021)Schubert, Voss, Cammarata, Goh, and Olah]{schubert2021high-low}
Ludwig Schubert, Chelsea Voss, Nick Cammarata, Gabriel Goh, and Chris Olah.
\newblock High-low frequency detectors.
\newblock \emph{Distill}, 2021.
\newblock \doi{10.23915/distill.00024.005}.
\newblock https://distill.pub/2020/circuits/frequency-edges.

\bibitem[Solomonoff(2009)]{Solomonoff2009}
Ray~J. Solomonoff.
\newblock \emph{Algorithmic Probability: Theory and Applications}, pp.\  1--23.
\newblock Springer US, Boston, MA, 2009.
\newblock ISBN 978-0-387-84816-7.
\newblock \doi{10.1007/978-0-387-84816-7_1}.
\newblock URL \url{https://doi.org/10.1007/978-0-387-84816-7_1}.

\bibitem[S{\o}nderby et~al.(2016)S{\o}nderby, Raiko, Maal{\o}e, S{\o}nderby, and Winther]{sonderby2016ladder}
Casper~Kaae S{\o}nderby, Tapani Raiko, Lars Maal{\o}e, S{\o}ren~Kaae S{\o}nderby, and Ole Winther.
\newblock Ladder variational autoencoders.
\newblock \emph{Advances in neural information processing systems}, 29, 2016.

\bibitem[Tan \& Le(2019)Tan and Le]{tan2019efficientnet}
Mingxing Tan and Quoc Le.
\newblock Efficientnet: Rethinking model scaling for convolutional neural networks.
\newblock In \emph{International conference on machine learning}, pp.\  6105--6114. PMLR, 2019.

\bibitem[Tibshirani(1996)]{51791361-8fe2-38d5-959f-ae8d048b490d}
Robert Tibshirani.
\newblock Regression shrinkage and selection via the lasso.
\newblock \emph{Journal of the Royal Statistical Society. Series B (Methodological)}, 58\penalty0 (1):\penalty0 267--288, 1996.
\newblock ISSN 00359246.
\newblock URL \url{http://www.jstor.org/stable/2346178}.

\bibitem[Touvron et~al.(2023)Touvron, Martin, Stone, Albert, Almahairi, Babaei, Bashlykov, Batra, Bhargava, Bhosale, et~al.]{touvron2023llama}
Hugo Touvron, Louis Martin, Kevin Stone, Peter Albert, Amjad Almahairi, Yasmine Babaei, Nikolay Bashlykov, Soumya Batra, Prajjwal Bhargava, Shruti Bhosale, et~al.
\newblock Llama 2: Open foundation and fine-tuned chat models.
\newblock \emph{arXiv preprint arXiv:2307.09288}, 2023.

\bibitem[Vahdat \& Kautz(2020)Vahdat and Kautz]{vahdat2020nvae}
Arash Vahdat and Jan Kautz.
\newblock Nvae: A deep hierarchical variational autoencoder.
\newblock \emph{Advances in neural information processing systems}, 33:\penalty0 19667--19679, 2020.

\bibitem[Vaswani et~al.(2017)Vaswani, Shazeer, Parmar, Uszkoreit, Jones, Gomez, Kaiser, and Polosukhin]{vaswani2017attention}
Ashish Vaswani, Noam Shazeer, Niki Parmar, Jakob Uszkoreit, Llion Jones, Aidan~N Gomez, {\L}ukasz Kaiser, and Illia Polosukhin.
\newblock Attention is all you need.
\newblock \emph{Advances in neural information processing systems}, 30, 2017.

\bibitem[Wang et~al.(2022)Wang, Variengien, Conmy, Shlegeris, and Steinhardt]{wang2022interpretability}
Kevin Wang, Alexandre Variengien, Arthur Conmy, Buck Shlegeris, and Jacob Steinhardt.
\newblock Interpretability in the wild: a circuit for indirect object identification in gpt-2 small.
\newblock \emph{arXiv preprint arXiv:2211.00593}, 2022.

\bibitem[Wei et~al.(2022)Wei, Tay, Bommasani, Raffel, Zoph, Borgeaud, Yogatama, Bosma, Zhou, Metzler, et~al.]{wei2022emergent}
Jason Wei, Yi~Tay, Rishi Bommasani, Colin Raffel, Barret Zoph, Sebastian Borgeaud, Dani Yogatama, Maarten Bosma, Denny Zhou, Donald Metzler, et~al.
\newblock Emergent abilities of large language models.
\newblock \emph{arXiv preprint arXiv:2206.07682}, 2022.

\bibitem[White et~al.(2023)White, Safari, Sukthanker, Ru, Elsken, Zela, Dey, and Hutter]{white2023neural}
Colin White, Mahmoud Safari, Rhea Sukthanker, Binxin Ru, Thomas Elsken, Arber Zela, Debadeepta Dey, and Frank Hutter.
\newblock Neural architecture search: Insights from 1000 papers.
\newblock \emph{arXiv preprint arXiv:2301.08727}, 2023.

\bibitem[Wu et~al.(2022)Wu, Cui, Pei, and Zhao]{GNNBook2022}
Lingfei Wu, Peng Cui, Jian Pei, and Liang Zhao.
\newblock \emph{Graph Neural Networks: Foundations, Frontiers, and Applications}.
\newblock Springer Singapore, Singapore, 2022.

\bibitem[Ying et~al.(2021)Ying, Cai, Luo, Zheng, Ke, He, Shen, and Liu]{ying2021transformers}
Chengxuan Ying, Tianle Cai, Shengjie Luo, Shuxin Zheng, Guolin Ke, Di~He, Yanming Shen, and Tie-Yan Liu.
\newblock Do transformers really perform badly for graph representation?
\newblock \emph{Advances in Neural Information Processing Systems}, 34:\penalty0 28877--28888, 2021.

\bibitem[Yuan et~al.(2023)Yuan, Yuan, Tan, Wang, and Huang]{yuan2023well}
Zheng Yuan, Hongyi Yuan, Chuanqi Tan, Wei Wang, and Songfang Huang.
\newblock How well do large language models perform in arithmetic tasks?
\newblock \emph{arXiv preprint arXiv:2304.02015}, 2023.

\bibitem[Zhang et~al.(2018)Zhang, Ren, and Urtasun]{zhang2018graph}
Chris Zhang, Mengye Ren, and Raquel Urtasun.
\newblock Graph hypernetworks for neural architecture search.
\newblock \emph{arXiv preprint arXiv:1810.05749}, 2018.

\bibitem[Zhong et~al.(2023)Zhong, Liu, Tegmark, and Andreas]{zhong2023clock}
Ziqian Zhong, Ziming Liu, Max Tegmark, and Jacob Andreas.
\newblock The clock and the pizza: Two stories in mechanistic explanation of neural networks.
\newblock \emph{arXiv preprint arXiv:2306.17844}, 2023.

\end{thebibliography}
\bibliographystyle{iclr2024_conference}
\newpage
\appendix

\section{Intuition of Why Hypernetworks Work}\label{app:hypernetwork-intuition}

In this section, we seek to provide a rough understanding of how we use our hypernetwork to generate weights and why we expect the resulting networks to be interpretable. Most important is the fact that the hypernetwork's outputs are used as the weights of the neural network \citep{chauhan2023brief}. Altogether, our hypernetwork architecture consists of two graph transformers \citep{ying2021transformers} which operate on the computation graph of the target network, serving as the encoder and decoder side of a hierarchical variational autoencoder (HVAE) \citep{vahdat2020nvae,sonderby2016ladder,child2020very}. The weights of the hypernetwork are generated by a Pareto hyperhypernetwork \citep{navon2020learning} which receives the HVAE $\beta$ hyperparameter \citep{higgins2017betavae} as input. The exact details of our hypernetwork and hyperhypernetwork architectures can be found in Appendix \ref{sec:attentional_hypernetworks}. Since our hypernetwork is a merge between a graph transformer and a HVAE, there are multiple lenses through which we understand our hypernetwork design:

\textbf{The Hypernetwork Interpretation.} The hypernetwork is a neural network whose output is used as the weights for a neural network \citep{chauhan2023brief}. For input, the hypernetwork is given information about every weight it needs to generate, such as the layer number and indices to identify the input and output neurons that it is connected to. The hypernetwork can then learn a general process for configuring each individual weight depending on its location within the network. Simpler configurations are easier to learn, and thus the resulting networks tend to be simpler, and thus more interpretable.

\textbf{The GNN Interpretation.} A neural network is a computation graph, so the most natural way to manipulate weight data is through a graph neural network (GNN) \citep{GNNBook2022,4700287,sanchez-lengeling2021a,daigavane2021understanding}. The hypernetwork is a graph transformer, whereby information is stored at the edges and is sent to and from adjacent nodes where attention heads operate. This allows the hypernetwork to compute based on how weights are mutually related to one another within the architecture (which itself does not need to be fixed either). This allows the hypernetwork to form structures of weights which are connected to the way the architecture is structured and are thus more likely to be interpretable.


\textbf{The MDL/Compression Interpretation.} Our hypernetwork is an application of the minimum description length (MDL) principle, which treats learning as a data compression procedure described by a mathematical version of Occam's razor \citep{mdlbook,grunwald2019minimum,RISSANEN1978465,Solomonoff2009}. The MDL principle treats a neural network as a compressor, so that we can speak of the “Occam simplicity'' of a model through KL divergences, which measure how well that model compresses a dataset in an information theoretic sense. \citep{10.1145/321892.321894,polyanskiytextbook} For our case, the dataset consists of the neural network weights and the compressor is the hypernetwork; the hypernetwork is an encoding/decoding system for compressing neural network weights into as simple a latent representation as possible, and networks derived from simpler representations are more interpretable.

\textbf{The Generative Model Interpretation.} Our hypernetwork is a generative model for data in the form of neural network weights. In the past, generative models were developed for and have been successful in generating human-interpretable forms of data, such as natural language \citep{openai2023gpt4,touvron2023llama} and images \citep{ho2020denoising}. Thus, we may expect that a generative model for data in the form of neural network weights would naturally have an inductive bias for human-interpretable weights.

\section{Method}

\subsection{Force-Directed Graph Drawings} \label{sec:graph_drawing}

Throughout this paper, we use a force-directed graph drawing algorithm to visualize neural networks as computation graphs. Force-directed graph drawing algorithms try to position every node in the plane to minimize clutter and account for several visual quality measures, such as edge overlaps, drawing area, and symmetry. Typically, such an algorithm defines an energy consisting of a sum of several components which each depend on the node positions, and the node positions are adjusted to minimize this energy via gradient descent. In our case, we apply three components: $1/r$ pairwise repulsion between all neurons, $kr^2$ pairwise attraction for weights of absolute value $k$, and $r^2$ attraction pulling all neurons towards the origin.

This graph drawing algorithm will help us to observe the structure of weights in the neural networks which we train in this paper, so that we may more easily understand how they function. Most importantly, it allows us to observe modularity, which is when individual parts of a learned neural network compute their operations individually without interaction. This is because the modules may form separate connected components which stay self-connected but repel each other in the drawn graph, making components easy to identify.

One of the main issues with force-directed graph drawing is that the gradient descent often falls into local minima of the energy, since two separate modules in the network can become tangled up, after which point the modules can no longer slide past one another and separate properly. To remedy this, we position the nodes in four dimensions instead of two (this provides mode connectivity), and we apply an increasingly strong decay to the two excess dimensions during gradient descent until they disappear, leaving a fully two-dimensional arrangement. This arrangement can be rescaled as needed for the visualization.

\subsection{Attentional Hypernetworks} \label{sec:attentional_hypernetworks}

\begin{figure}[htbp]
    \centering
    \includegraphics[width=\textwidth]{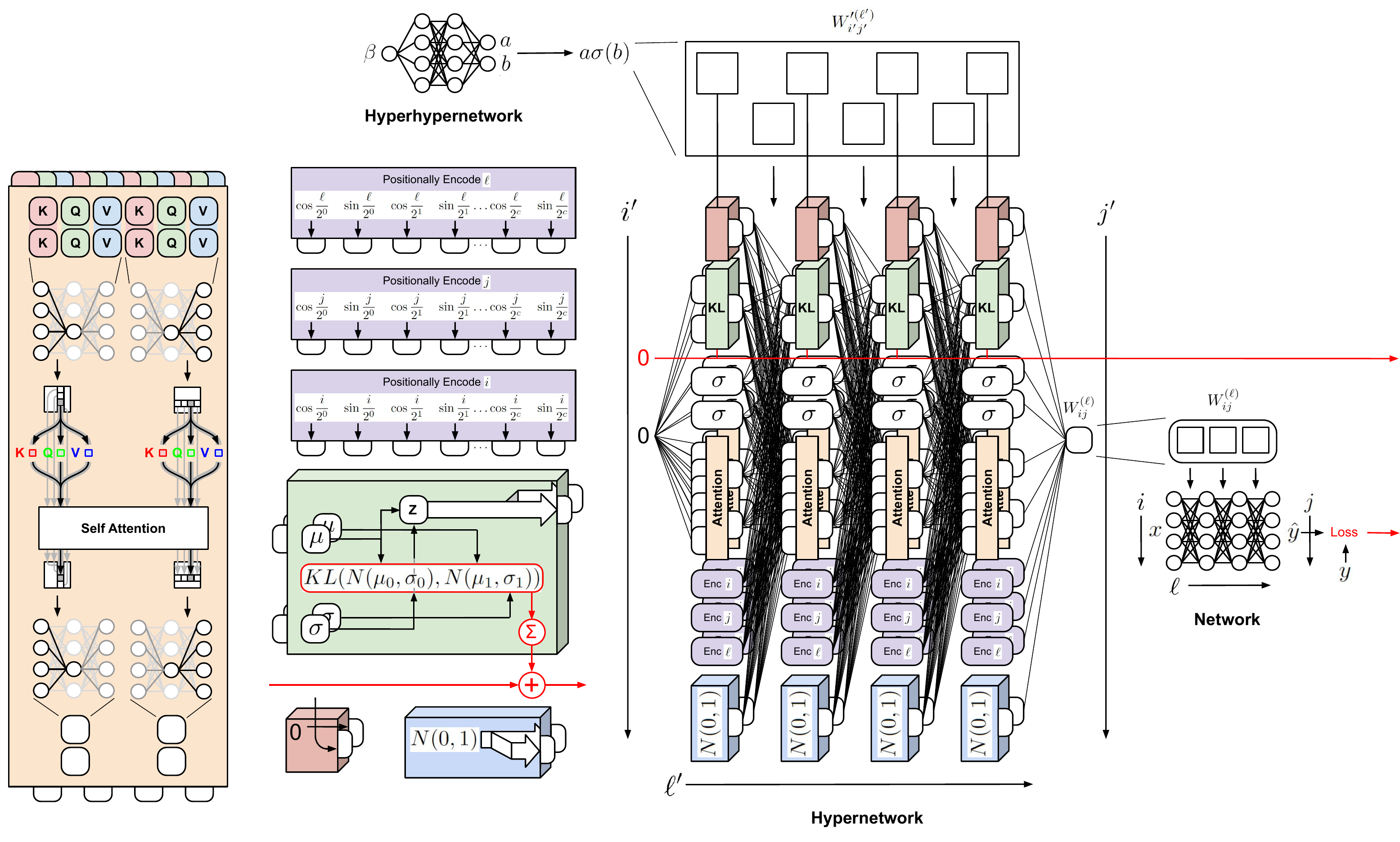}
    \caption{The full architecture of the hypernetwork. The hyperhypernetwork above generates weights for the hypernetwork, which generates weights for the network, on which the loss is evaluated. There are many components in the hypernetwork, each drawn individually on the left: graph attention, positional encodings, information bottleneck channels, learned hyperfeatures, and random variables.}
    \label{fig:architecture_overview}
\end{figure}

In this section, we explain in more detail how we use a hypernetwork to generate the weights of the MLP which we would like to train. Instead of learning the MLP weights directly, we learn the “hyperweights'' of this hypernetwork. We generate the MLP weights every iteration as part of the forward pass when doing gradient descent. Figure \ref{fig:architecture_overview} fully depicts all the components of the hypernetwork, and in the following text and sections below, we will explore each of the components in detail.

We design the hypernetwork in a way so that it has an inductive bias to create certain structures and formations of weights with more ease than others. For example, we may want the hypernetwork to tell the weights how to self-organize into many duplicates of a specific circuit, which are then connected together in a formation, rather than many unrelated circuits connected in a more complex manner. Given that these structures are organized, with a limited number of hyperweights, it might be easier to learn a method of constructing such structures rather than the structures themselves. We believe that and inductive bias for learning such methods should best arise from clever forms of parameter reuse, just like how convolutional filters are best for translationally equivariant computations. As such, our hypernetwork operates much like a graph transformer, whose computations are duplicated across all nodes and edges.

For every component of the MLP, there is an analogous component for our hypernetwork, which itself can somewhat be thought of like an MLP. For example, the hypernetwork has hyperfeatures, hyperlayers, hyperwidth, hyperdepth, and hyperactivations in the same way that the MLP (or “network”) has features, layers, width, depth, and activations.

For our network architecture, we restrict ourselves to a neural network consisting of weights $W^{(\ell)} \in \mathbb{R}^{n_{i+1} \times n_i}$ and biases $b^{(\ell)} \in \mathbb{R}^{n_{i+1}}$ with $N$ layers:
\begin{align*}
a^{(\ell+1)}_j = \sigma(b^{(\ell)}_j + \sum_{i} W^{(\ell)}_{ij}a^{(\ell)}_i)
\end{align*}
with $a^{(1)}_i$ the input vector and $b^{(N)}_j + \sum_{i} W^{(N)}_{ij}a^{(N)}_i$ the output vector.

We will now describe our hypernetwork architecture. The fundamental unit of data processed by the hypernetwork is a high-dimensional ragged tensor of “hyperactivations'' containing information about every weight in the network. Each “hyperlayer'' operates by applying many operations which serve to perform computations that only conduct information along individual axes of the tensor. For example, if we have a hyperactivation tensor pertaining to a 4 dimensional CNN filter tensor with an $x$ axis, we might apply a blur filter along the corresponding $x$ axis of the hyperactivation tensor while treating all other dimensions as batch dimensions, duplicating this operation for all indices.

Returning from the CNN example to our MLP network, we designate the last axis as special, and we call it the “hyperfeature'' dimension, analogous to the feature dimension in the MLP. The hyperactivation tensor is sliced into many blocks along the hyperfeature axis, and each block undergoes its own operation along its own designated axis (all other axes treated as batch dimensions), then the results are concatenated back together in the hyperfeature dimension. A linear operation in the hyperfeature axis can then be used to control the movement of data between these individual blocks, allowing for movement of data in every direction of the tensor. A good analogy is the way that trains can be shunted onto different tracks, where they may travel in different directions or undergo different procedures. This all can then be repeated multiple times by stacking together multiple of these hyperlayers. We will index hyperlayers using the variable $\ell'$ since we are indexing layers with $\ell$, and let us call the number of hyperlayers the “hyperdepth'' $N'$, which we set to 4.

We will now roughly describe how we construct a hyperactivation tensor from the MLP weights. We can summarize the MLP weights using a ragged tensor $W_{ij}^{(\ell)}$ indexed by input neuron $i$, output neuron $j$, and layer $\ell$. The hyperactivation tensor then has the same shape, except that it has an additional hyperfeature axis indexed by a variable $i'$.

The operations performed on blocks of the hyperactivation tensor are treated like activation functions for the hypernetwork. For a hyperactivation tensor with indices $i$, $j$, $\ell$, and $i'$, a block is processed with each of the following operations:
\begin{itemize}
    \item $x \rightarrow \sigma(x)$. Elementwise activation function. Used on a block of 20 hyperfeatures.
    \item Nothing $\rightarrow$ Positional encoding of $i$ if $\ell=1$, else a zero tensor.
    \item Nothing $\rightarrow$ Positional encoding of $j$ if $\ell=N$, else a zero tensor.
    \item Nothing $\rightarrow$ Positional encoding of $\ell$.
    \item Nothing $\rightarrow$ 5 hyperfeatures of random i.i.d. samples from a standard normal distribution.
    \item A self-attention head for every neuron in the network. Each edge feeds 3 sub-blocks of the hyperactivation tensor---the keys, queries, and values---to the neuron in front and another 3 sub-blocks to the neuron behind, and concatenates the results received from both sides. The keys, queries, and values are all 5 hyperfeatures in size.
\end{itemize}

The output of the final hyperlayer has two hyperfeatures---one is used as the weight tensor and the other is averaged along the input neuron axis $i$ and is used as the bias tensor.

Notice that this hypernetwork architecture as it stands does not contain that many learned parameters, compared to how many MLP parameters it can generate. Even with so few parameters, it still takes a huge amount of computation, which is necessary for the amount of parameter reuse we want. Remember that parameter reuse is useful for the hypernetwork to produce highly patterned structures in the network's weights.


\subsubsection{KL Attentional Hypernetworks}

Plain attentional hypernetworks like the one described above have a certain defect: there might be some networks that are learned more easily by gradient descent, but which are hard for the hypernetwork to capture because they are not structured in any obvious fashion. To allow the attentional hypernetwork to capture these circuits, we introduce another axis to the hyperactivation tensor, which we call the KL axis.

The KL axis has two indices, representing the encoder and decoder side of an information channel. Information is passed through this axis via two operations:
\begin{itemize}
    \item Nothing $\rightarrow$ 5 hyperfeatures of learned variables which go along with the hyperweights during training. These hyperfeatures are intended to capture information that can be learned by standard gradient descent, but they are only given on the encoder side and are set to zero for the decoder side.
    \item Start with two blocks of 4 hyperfeatures representing $\mu$ and $\sigma$ values for normal distributions. The total KL divergence $D_{\text{KL}}(q||p)$ between the encoder-side and decoder-side distributions $q$ and $p$ is computed and added to an accumulator variable. The output is 4 hyperfeatures of samples from $q$; a copy of these samples for each of the encoder and decoder sides to use.
\end{itemize}

The weights and biases are constructed using only the output from the decoder side.\footnote{By introducing the KL axis, we have essentially turned our hypernetwork into a kind of conditional hierarchical VAE.}

In this additional axis, learning via standard gradient descent is allowed to take place, since the hypernetwork can pass the learned variables provided to the encoder through the KL channel to the decoder, where the variables can be output to be treated as weights and biases. But notice that any information passing through the KL channel has its information content measured and accounted for in an accumulator variable. This means we can suppress the usage of gradient descent algorithms by regularizing the quantity of raw weight information which passes through. \footnote{Note that the feedback in the channel lets the hypernetwork perform simpler computations without destroying the ELBO bound in the VAE interpretation of our hypernetwork.} Tuning the suppression factor $\beta$ allows us to control the balance between a hypernetwork which only designs very structured patterns of weights and a hypernetwork which regurgitates unstructured patterns of learned and memorized weights. The balance between loss and KL implicitly encourages the neural network to develop structures like modules and duplicated circuits of neurons, as these are structures that the hypernetwork can encode in a more compressed manner and will be penalized less for.

\subsubsection{Hyperhypernetworks for Multi-Objective Optimization}

We are now left with a multi-objective optimization problem where we would like to jointly optimize for the network's loss and hypernetwork's total accumulated KL. To solve this, we use a hyperhypernetwork to generate the hyperweights in every step during the forward pass, using something like a Pareto Hypernetwork. This hyperhypernetwork takes $\log \beta$ (rescaled and shifted) as input, has two hidden layers of size 100 and 10 with swish activation, and outputs two vectors $a$ and $b$ where $a\sigma(b)$ is treated as the flattened vector of hyperweights. At every iteration, we sample $\beta$ from a distribution, use the hyperhypernetwork to generate the hyperweights, use the hypernetwork to generate the weights, and use $\log (L + \beta D_\text{KL})$ as the objective, where $L$ is the network's loss and $D_\text{KL}$ is the hypernetwork's accumulated KL.

\end{document}